\documentclass[letterpaper, 10 pt, conference]{ieeeconf}  % Comment this line out if you need a4paper

\IEEEoverridecommandlockouts                              % This command is only needed if 
\usepackage{cite}
\usepackage{amsmath,amssymb,amsfonts}
\usepackage{algorithmic}
\usepackage{graphicx}
\usepackage{textcomp}
\usepackage{xcolor}

\usepackage[font=footnotesize]{caption}
\usepackage[font=footnotesize]{subcaption}
\usepackage{hyperref}
\usepackage{tikz}
\newcommand\copyrighttext{%
  \footnotesize \textcopyright\space2025 IEEE. Personal use of this material is permitted. Permission from IEEE must be obtained for all other uses, in any current or future media, including reprinting/republishing this material for advertising or promotional purposes, creating new collective works, for resale or redistribution to servers or lists, or reuse of any copyrighted component of this work in other works.}
\newcommand\copyrightnotice{%
\begin{tikzpicture}[remember picture,overlay]
\node[anchor=south,yshift=10pt] at (current page.south) {\fbox{\parbox{\dimexpr\textwidth-\fboxsep-\fboxrule\relax}{\copyrighttext}}};
\end{tikzpicture}%
}

\title{\LARGE \bf
Leader-follower formation enabled by pressure sensing in free-swimming undulatory robotic fish
}

\author{Kundan Panta\textsuperscript{1}, Hankun Deng\textsuperscript{1}, Micah DeLattre\textsuperscript{1}, and Bo Cheng\textsuperscript{1*}, \textit{Member, IEEE}% <-this % stops a space
\thanks{Research supported by the National Science Foundation [NSF award no. CNS-2334881, awarded to B.C.] and the Army Research Office [ARO grant no. W911NF-20-1-0226, awarded to B.C.].}% <-this % stops a space
\thanks{\textsuperscript{1}Department of Mechanical Engineering, The Pennsylvania State University, University Park, PA 16802, USA.}%
\thanks{\textsuperscript{*}Corresponding author: \href{mailto:buc10@psu.edu}{buc10@psu.edu}.}%
}

\begin{document}
% \title{Leader-follower formation in free-swimming undulatory robotic fish enabled by pressure sensing
% \thanks{Research supported by the National Science Foundation [NSF award no. CNS-2334881, awarded to B.C.] and the Army Research Office [ARO grant no. W911NF-20-1-0226, awarded to B.C.].
% \textsuperscript{1}Department of Mechanical Engineering, The Pennsylvania State University, University Park, PA 16802, USA.
% \textsuperscript{*}Corresponding author: \href{mailto:buc10@psu.edu}{buc10@psu.edu}.}
% }
% \author{Kundan Panta\textsuperscript{1}, Hankun Deng\textsuperscript{1}, Micah DeLattre\textsuperscript{1}, and Bo Cheng\textsuperscript{1*}, \textit{Member, IEEE}}
\maketitle
\copyrightnotice
% \thispagestyle{empty}
% \pagestyle{empty}
%%%%%%%%%%%%%%%%%%%%%%%%% Abstract %%%%%%%%%%%%%%%%%%%%%%%%%
\begin{abstract}
Fish use their lateral lines to sense flows and pressure gradients, enabling them to detect nearby objects and organisms. Towards replicating this capability, we demonstrated successful leader-follower formation swimming using flow pressure sensing in our undulatory robotic fish (µBot/MUBot). The follower µBot is equipped at its head with bilateral pressure sensors to detect signals excited by both its own and the leader's movements. First, using experiments with static formations between an undulating leader and a stationary follower, we determined the formation that resulted in strong pressure variations measured by the follower. This formation was then selected as the desired formation in free swimming for obtaining an expert policy. Next, a long short-term memory neural network was used as the control policy that maps the pressure signals along with the robot motor commands and the Euler angles (measured by the onboard IMU) to the steering command. The policy was trained to imitate the expert policy using behavior cloning and Dataset Aggregation (DAgger). The results show that with merely two bilateral pressure sensors and less than one hour of training data, the follower effectively tracked the leader within distances of up to 200 mm (= 1 body length) while swimming at speeds of 155 mm/s (= 0.8 body lengths/s). This work highlights the potential of fish-inspired robots to effectively navigate fluid environments and achieve formation swimming through the use of flow pressure feedback.

\textit{Video}---\url{https://youtu.be/DIDYGi9Td0I}
\end{abstract}

% \begin{IEEEkeywords}
% artificial lateral line, fish robot, schooling, imitation learning
% \end{IEEEkeywords}

%%%%%%%%%%%%%%%%%%%%%%%%% Intro %%%%%%%%%%%%%%%%%%%%%%%%%
\section{Introduction}
Fish can adapt their swimming gaits to exploit existing flow structures in the environment or those generated by other fishes to gain hydrodynamic benefits \cite{liao_review_2007, liao_fish_2003, thandiackal_-line_2022, zhang_collective_2024}. They can also leverage information encoded in flow structures for perception, especially at short ranges. For instance, fish use superficial and canal neuromasts to detect surface shear flow and pressure gradients (e.g., generated by predators and prey\cite{engelmann_hydrodynamic_2000}) along their lateral line \cite{Bleckmann2009, mogdans_fish_2003}, and swim in wakes for schooling and station-keeping, even without visual cues \cite{ko_role_2023, liao_role_2006, partridge_sensory_1980, Pitcher1976}.

%or instance, fish detect flow disturbances from predators and prey using their lateral lines

%However, because sensory inputs are coupled with body morphology and control, replicating lateral line sensing on rigid underwater vehicles \cite{Xu2017, yen_localization_2020} may not provide the same hydrodynamic information available to natural, undulating swimmers \cite{zhang_lateral_2015, qiu_locating_2023, zheng_artificial_2019}. 

\begin{figure}[th]
    \centering
    \includegraphics[]{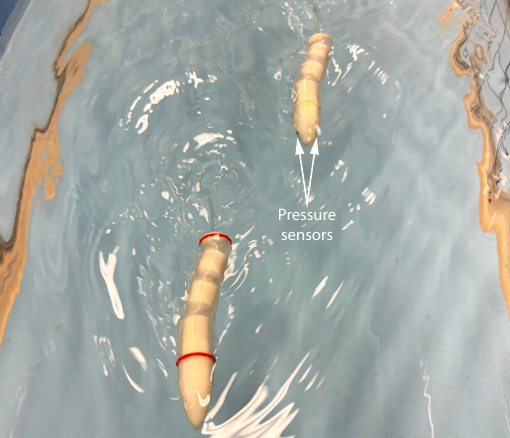}
    \caption{The follower µBot steers towards the leader, informed by its pressure sensors and IMU, which encode the hydrodynamic interactions between the two µBots.}
    \label{fig:highlight}
\end{figure}

Robotic fish typically mimic this flow feedback using off-the-shelf static pressure sensors placed near the head \cite{Kruusmaa2014, yen_localization_2020, qiu_locating_2023, zheng_artificial_2019}, though alternative mechanisms exist \cite{Thandiackal2021, DeVries2015, jiang_underwater_2022}. However, pressure sensors are often affected by static pressure changes, such as water splashes or body roll, which can obscure hydrodynamic signals essential for environmental perception \cite{Paez2021, qiu_locating_2023}. Despite these limitations, it may be feasible to detect relevant flow stimuli using pressure sensor arrays and signal processing techniques \cite{Kruusmaa2014, yen_localization_2020, Zheng2021a}. To advance the flow-based perception and control capability in robotic swimmers, here we aim to achieve leader-follower formation swimming using flow
pressure sensing in our undulatory robotic fish (i.e., µBots \cite{deng_development_2023}) (Fig. \ref{fig:highlight}).
%Therefore this work marks a critical step forward towards intelligent, hydrodynamically-informed interactions with its environment.

With onboard pressure feedback, a follower µBot could theoretically detect the flow generated by a nearby leader µBot by decoding the pressure information and maintain a leader-follower formation \cite{zheng_artificial_2019}. However, this task is challenging due to the complex nonlinear hydrodynamic interactions between the µBots. Previous studies have explored how underwater robots can detect the position and orientation of a leading robot or other flow sources, though these robots often have rigid, non-undulating bodies \cite{yen_localization_2020, Xu2017, xie_artificial_2017, Zheng2021a}. Importantly, in addition to the flow generated by the leader, the follower’s own undulating movements affect the flow feedback \cite{zheng_artificial_2019}, which can hinder the detection of external objects \cite{windsor_swimming_2008}. Some studies on hydrodynamic feedback in oscillating robotic fish have simplified the problem by limiting the robot’s degrees of freedom \cite{zheng_artificial_2019} or using a fixed, simpler flow source \cite{qiu_locating_2023}.
%Differing from these setups, the hydrodynamic interactions both shape and are influenced by the propulsion and relative states between leader and follower \cite{Verma2018, Li2020, liao_fish_2022}. 
Therefore, enabling leader-follower formation in free-swimming, undulating µBots presents novel challenges, as the follower must distinguish the pressure variations generated by the leader from those due to ego motion, both of which vary dynamically. In addition, hydrodynamic pressure feedback is typically noisy and available at limited locations on the robot body (due to the practical challenge of integrating sensors in small robotic swimmers such as µBots). Therefore, it is crucial to extract information from the noisy temporal pressure patterns with these limited pressure channels \cite{Verma2018, qiu_locating_2023, panta_touchless_2024}. 

%In addition to the various sources of excitation for pressure sensors, only a limited number can be practically integrated onto the follower to observe the high-dimensional fluid-robot system. 

Imitation learning offers a goal-directed framework to solve the above challenges and train a leader-following control policy. If an expert policy---potentially with access to the leader’s true positions and orientations, which are normally unavailable to the follower---can be provided, a supervised learning problem can be framed \cite{ross_reduction_2011, pomerleau_alvinn_1988}, mapping pressure signals to the expert's actions at each time step. In this scenario, recurrent neural networks, particularly those with long short-term memory (LSTM) units \cite{Hochreiter1997}, are effective as they can extract information from the input history. Imitation learning, combined with LSTM, may provide a practical way to train an end-to-end mapping from instantaneous pressure  measurements to control (steering) actions, offering a model-free alternative to rigorous analysis of the imperfect pressure signals \cite{qiu_locating_2023, Zheng2021a}.

Therefore, this work aims to
%\begin{enumerate}[leftmargin=*]
%\item 
demonstrate, for the first time in experiments (to the best of the authors' knowledge), that a follower µBot can track a randomly moving leader using pressure feedback in free-swimming, undulating robotic fish.
%\item  
Towards this goal, a fully data-driven controller through imitation learning is developed. %\end{enumerate}

%%%%%%%%%%%%%%%%%%%%%%%%% Methods %%%%%%%%%%%%%%%%%%%%%%%%%
\section{Materials and Methods}
\begin{figure*}[ht]
    \centering
    
    \tabskip=0pt
    \valign{#\cr
      \hbox{%
        \begin{subfigure}{2.9in}
        \centering
        \includegraphics{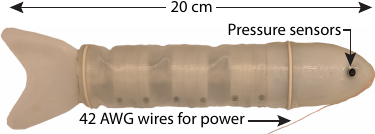}
        \caption{}
        \end{subfigure}%
      }\vfill
      \hbox{%
        \begin{subfigure}{2.9in}
        \centering
        \includegraphics{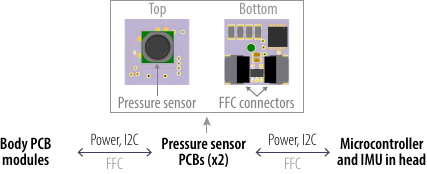}
        \caption{}
        \end{subfigure}%
      }\cr
      \noalign{\hfill}
      \hbox{%
        \begin{subfigure}[b]{4in}
        \centering
        \includegraphics{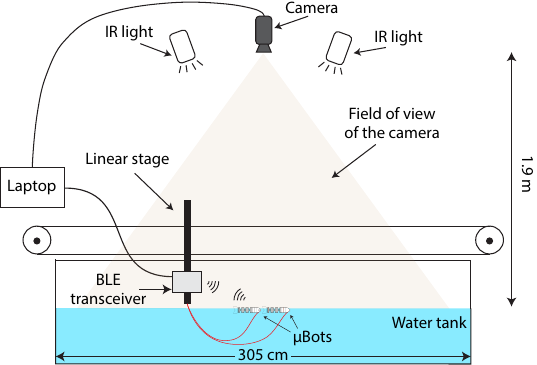}
        \caption{}
        \end{subfigure}%
      }\cr
    }
    
    \caption{(a) The follower µBot with a pressure sensor integrated on each side of its head module. (b) The modular PCBs onto which the pressure sensors are mounted, connected serially to the microcontroller and body modules via flat-flex cable (FFC) connectors. (c) The still water tank used for the leader-follower experiments.}
    \label{fig:hardware}
\end{figure*}

\subsection{Integration of Pressure Sensors}
To enable the follower µBot to perceive flow, we integrated off-the-shelf Honeywell MPRSL0001PG00001C pressure sensors into its head. These sensors were chosen for their low 6.9 kPa range, built-in 24-bit ADC, I2C communication, water compatibility, and compact size. We designed modular PCBs to mount the sensors, regulate voltage using the TPS71433DCKR (Texas Instruments), and configure I2C addresses with the LTC4316 (Analog Devices). This setup allowed us to connect the pressure sensors to the existing µBot using the same flat-flex cables, ensuring all sensors—including the inertial measurement unit (IMU) (Bosch BNO055)—and motor drivers could be accessed via the same I2C bus. While biological fish and other robotic fish often have a large array of sensors near their head, we first aimed to test the two bilateral sensors here (while a 6-sensor version has already been built and will be tested in future work).
%and assess perception by comparing pressure differences between the left and right sides.

%Holes were added to the follower's head module to expose the pressure sensor ports, and the gaps around the sensors were filled with silicone rubber (Ecoflex 00-35) for waterproofing.
% Cite Liao for concentration of neuromasts on fish head.
		
To improve experimental feasibility and consistency, we replaced the batteries powering the existing µBots with an external power supply to enable longer experiments and maintain consistent voltage levels. 
Three 42 AWG magnet wires were carried on a linear stage to power the electronics and actuators at 15 V; these thin, lightweight wires had minimal impact on propulsion \cite{Deng2021} and were routed downward into the water, trailing behind the µBots to avoid affecting hydrodynamic interactions between the leader and follower. Notably, we can revert to battery power once continuous data collection is no longer needed. 
%Second, we resumed manufacturing the µBot's rubber suits by curing silicone rubber (Ecoflex 00-20) in molds, similar to the process used for the gen 1 µBot. This approach provided greater precision and consistency in suit thickness, leading to more uniform swimming performance between the two µBots. Additionally, we manufactured the suit in one piece rather than per segment and added internal protrusions to align it with the grooves on each body segment.These adjustments ensured consistent tension across the suit and eliminated the need for O-rings on intermediate segments, improving both assembly consistency and ease.
% Justify the thin wires not affecting the hydrodynamic interactions.

\subsection{Experimental Setup}
The experiments were conducted in a water tank (305 cm L $\times$ 58 cm W$\times$ 56 cm H) (Fig. \ref{fig:hardware}). 
%We used a linear stage to carry the wires powering the µBots, keeping them free of tension. However, in this case, we manually moved the stage, staying just behind the swimming µBots. 
The overall setup was modified from that used previously for path-following by µBot \cite{deng_development_2023}. We updated our image processing routines to track and low-pass filter the 2D positions and headings of both µBots in real-time using an overhead infrared camera (Basler acA2000-165umNIR). The µBots continued to communicate with a host laptop via Bluetooth Low Energy (BLE), with the host receiving sensor readings and motor command data and sending new motor commands to the µBots' microcontrollers. 
%To minimize communication distance, we mounted USB BLE dongles (DFRobot Bluno Link) on the linear stage. 
The one-way communication latency averaged 60-80 ms for data transmission. All image and data processing occurred in real-time at 50 Hz using MATLAB 2024a.

\subsection{Generating Forward Swimming and Steering Gaits}\label{sec:gaits}
We adopted the Kuramoto central pattern generator (CPG) to generate actuator torque commands for each robot joint, which has also been applied to other bio-inspired robots \cite{Bayiz2019, Ijspeert2007, Crespi2008, Crespi2008a}.
%In previous work, we used the Matsuoka central pattern generator (CPG)
The Kuramoto CPG allowed us to specify desired frequencies, amplitudes, offsets, and intersegmental phase offsets for the oscillating actuator torques. We tuned the CPG parameters to match the gaits from previous experiments \cite{deng_development_2023}, resulting in phase offsets of -65° between successive body segments (head to tail) at a frequency of 5 Hz. By adjusting the amplitude offset across the actuators, we could bend the µBot to steer in the desired direction. Torque commands exceeding the actuation limits due to steering were truncated accordingly. To prevent excessive bending, which reduces thrust, the steering offset was constrained to within \textpm30\% of the oscillation amplitude. For simplicity, we fixed all other CPG parameters and made the steering offset the sole control variable.

\subsection{Random Path Generation for the Leader}\label{sec:leader_path}
\begin{figure}
    \centering
    \includegraphics{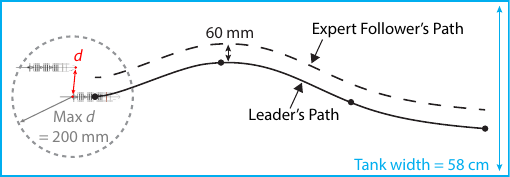}
    \caption{Generation of the leader's random path and the expert follower's staggered path. The maximum allowable distance between the leader's tail and the follower's nose, $d$, during leader-following is capped at 200 mm.}
    \label{fig:paths}
\end{figure}

To generate random paths for the leader, we fitted a spline through its initial position and three randomly placed control points evenly distributed along the length of the tank (see an example in Fig. \ref{fig:paths}). We then applied the same method from our previous work \cite{deng_development_2023}, using line-of-sight (LOS) guidance \cite{Kelasidi2016} to determine the reference heading angle and generating steering commands proportional to the heading error.

\subsection{Imitation Learning}
\subsubsection{Expert Policy}\label{sec:expert}
\begin{figure*}
    \centering
    \includegraphics{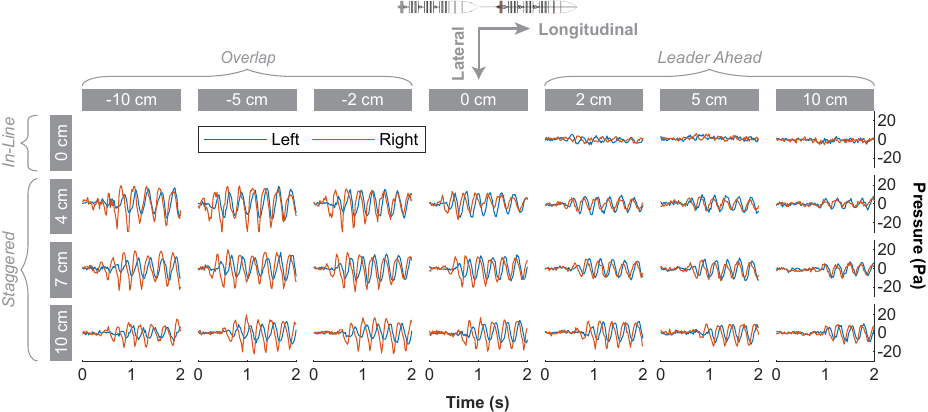}
    \caption{Pressure signals at various lateral and longitudinal distances between a fixed follower µBot and a free-swimming leader µBot moving from rest. The distances refer to the \emph{initial} distance between the leader's tail and the follower's nose. The leader was to the follower's right in the staggered formations.}
    \label{fig:fixed_follower}
\end{figure*}

To inform the expert policy, we first collected preliminary pressure data with the follower fixed while the free-swimming leader undulated at different positions relative to the follower (Fig. \ref{fig:fixed_follower}). The pressure variations were minimal when the follower was directly in line with the leader; however, they became more pronounced when the leader was staggered, with its caudal fin exciting the flow near the sensors—even up to 10 cm (= 1/2 body length) laterally. These effects lessened as the leader moved ahead (longitudinal distance\textgreater 0), and the delay in pressure changes increased with lateral distance. 
%Identical patterns were observed when the leader was on the left.
This suggests the follower can perceive the leader’s lateral and longitudinal distances from pressure cues within 10 cm, especially when staggered.

We designed the expert policy to take advantage of stronger pressure feedback in staggered leader-follower configurations, helping the follower learn reliable associations between pressure signals and steering commands. Using LOS guidance based on external motion tracking, the expert followed a staggered path generated by offsetting the leader's random path by 60 mm (= 0.3 body lengths, within the tested 10 cm range) toward the follower's initial position. Steering commands were calculated as described in Sec. \ref{sec:leader_path}. This expert policy allowed the follower to maintain a staggered configuration, providing rich flow information.

\subsubsection{Learner Policy}\label{sec:imitating_policy}
Unlike the expert, the learner policy did not have access to motion tracking data. Its inputs came from onboard sensors—two pressure signals, three Euler angles from the IMU in the head segment, and the last motor commands to the actuators. We included Euler angles and actuator commands to help the policy implicitly distinguish between self-generated and leader-induced pressure, without explicit pre-processing. The biases in the pressure and the yaw angle were removed before the µBots began moving in each experiment. The policy's output was the steering offset for leader-following.

We implemented the learner policy using a recurrent neural network, consisting of an LSTM layer to extract temporal patterns from the pressure inputs, a hidden fully-connected layer, dropout layers for regularization, and a fully-connected output layer with a tanh activation followed by a scaling layer to constrain the steering outputs. Each hidden layer has 64 units, selected in preliminary experiments for lower validation loss, resulting in a total of 22,912 parameters.

\subsubsection{Behavior Cloning}
\begin{figure}
    \centering
    \begin{subfigure}{0.46\linewidth}
        \centering
        \includegraphics{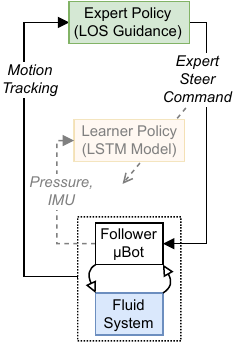}
        \caption{}
        \label{fig:flowchart_bc}
    \end{subfigure}
    \hfill
    \begin{subfigure}{0.46\linewidth}
        \centering
        \includegraphics{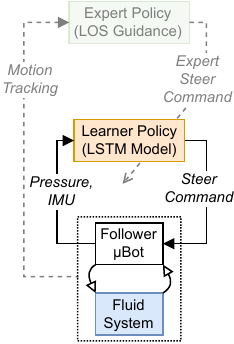}
        \caption{}
        \label{fig:flowchart_dagger}
    \end{subfigure}
    \caption{The two imitation learning approaches: (a) Behavior Cloning (BC) and (b) DAtaset Aggregation (DAgger).}
    \label{fig:flowcharts_imitation}
\end{figure}

In basic behavior cloning (BC), the expert provides demonstrations, and the observed states are mapped to the expert's actions. We began with this approach (Fig. \ref{fig:flowchart_bc}) to develop an initial learner policy. Using the expert's LOS guidance, we steered the follower near the randomly moving leader to collect 120 demonstrations, while recording the learner policy's inputs—onboard sensor readings and motor commands—in the background. We then applied the method in Sec. \ref{sec:imitating_policy} to map these inputs to the expert's steering commands.

\subsubsection{DAgger} Since the learner policy, which relies solely on pressure feedback, is likely to perform worse than the expert policy, which has access to the µBots' global positions, the follower can often end up in unfamiliar, sub-optimal states. To address this, we used DAgger (DAtaset Aggregation) \cite{ross_reduction_2011} to improve upon BC (Fig. \ref{fig:flowcharts_imitation}). In DAgger, we rolled out the learner policy, closing the loop between onboard sensors and steering commands via the LSTM model, while also querying the expert's LOS guidance at each time step for steering commands. After 20 rollouts, we split the data into training and validation sets, appended it to the BC data, and retrained the policy to imitate the expert's actions. This process was repeated until a satisfactory leader-following policy was learned.

\subsubsection{Training the Learner}
For supervised learning of the learner policies, we centered all input and output data and normalized them by their standard deviations. We used 90\% of the data for training and 10\% for validation in each learning iteration. The model was trained with the Adam optimizer for 10,000 epochs at a base learning rate of 0.005, and the weights from the epoch with the lowest validation loss were selected to mitigate overfitting.

\subsubsection{Experimental Methodology}\label{sec:exp_method}
In each rollout, the µBots started on one side of the tank, swam forward, and were manually returned to their starting positions after each rollout. The leader began 0.4 s (two undulation cycles) before the follower. Rollouts ended when one of three conditions was met: the follower's nose moved more than 200 mm (= 1 body length) from the leader's tail (Fig. \ref{fig:paths}), the follower touched the leader, or 10 seconds passed without either condition being violated. We had an equal number of rollouts with the leader on the left and on the right of the follower in all learning iterations. We started each rollout with both µBots parallel, the follower’s nose next to the leader’s last two segments, and a lateral separation of about 60 mm.
% The initial longitudinal overlap, lateral separation, and relative heading were \_\textpm\_ mm, \_\textpm\_ mm, and \_°\textpm\_°, respectively.

\subsection{Evaluating Leader-Following Performance}
\begin{figure}
    \centering
    \begin{subfigure}{0.38\linewidth}
        \centering
        \includegraphics{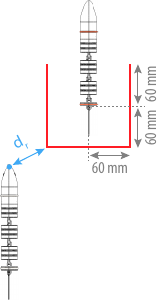}
        \caption{}
        \label{fig:reward_spatial}
    \end{subfigure}
    \begin{subfigure}{0.48\linewidth}
        \centering
        \includegraphics{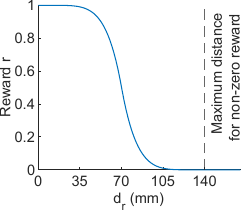}
        \caption{}
        \label{fig:reward_dist}
    \end{subfigure}
    \caption{(a) Definition of $d_r$ as the distance between the follower's nose and the square-shaped outline centered on the leader's tail segment at each time step. The square outline was chosen over a similar-sized circular one to keep distance between the maximum reward region and the leader's flapping tail. (b) Reward $r$ as a function of $d_{r}$, as defined by Eq. \ref{eq:reward}. This function was chosen to gradually map the reward from $d_r\in [0,140]$ to $r\in[0,1]$.}
    \label{fig:reward}
\end{figure}

We used two metrics to compare the learner policies with the expert policy and a no-steering policy, representing the upper and lower bounds of leader-following performance, respectively. The first metric was the mean absolute error per rollout between the learner and expert policies, where a lower error indicates better imitation. The second metric was a reward based on how well the follower maintained a 60 mm distance from the leader, which both the learner and expert policies aimed to achieve. The reward $r$ (Fig. \ref{fig:reward}) was: 
\begin{equation}
r=
    \begin{cases}
      1-f(d_{r}/140) & d_{r} \leq 70 \\
      f(1-d_{r}/140) & 70 < d_{r} \leq 140 \\
      0 & d_{r} > 140
   \end{cases}
\label{eq:reward}
\end{equation}
where $d_{r}$ is the shortest distance between the follower's nose and the outline centered at the leader's tail in Fig. \ref{fig:reward_spatial}, and
\begin{align*}
    s&=0.7,\\
    c&=2/(1-s)-1,\\
    f(x) &= (2x)^c/2
\end{align*}
where $s$ controls the curve's shape. The maximum reward per time step was 1, with a minimum of 0 when $d_{r}$\textgreater140 mm. The cumulative $r$ per rollout served as the second metric, with a maximum value of 500 (in 10 s = 500 time steps). Early terminations, due to the follower moving more than 200 mm away or touching the leader, led to lower values.

%%%%%%%%%%%%%%%%%%%%%%%%% Results %%%%%%%%%%%%%%%%%%%%%%%%%
\section{Results}
\begin{figure}
    \centering
    \begin{subfigure}{\linewidth}
        \centering
        \includegraphics{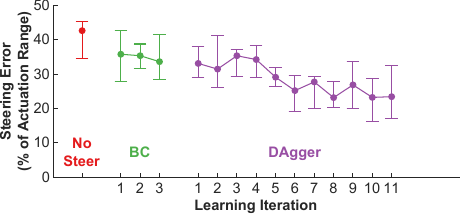}
        \caption{}
        \label{fig:results_steering_error_learning}
    \end{subfigure}
    \begin{subfigure}{\linewidth}
        \centering
        \includegraphics{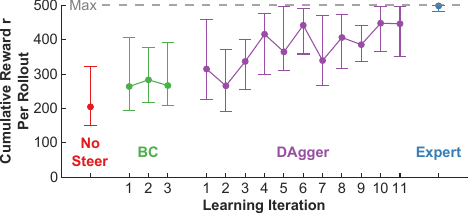}
        \caption{}
        \label{fig:results_rewards_learning}
    \end{subfigure}
    \caption{Trends in the metrics across learning iterations, showing the median and the 25\textsuperscript{th} and 75\textsuperscript{th} quartiles. (a) Mean absolute error in steering per rollout between the policy and the expert. (b) Cumulative reward $r$ per rollout.}
    \label{fig:results_metrics_learning}
\end{figure}

In the BC phase, three policies were trained with data from 40, 80, and 120 rollouts to evaluate how the performance scaled with the number of expert demonstrations. With 40 rollouts, the BC policies performed slightly better than the no-steering group on both metrics (Fig. \ref{fig:results_metrics_learning}), but additional demonstrations did not lead to further improvements. Significant improvements were observed in the DAgger phase, with the learner policy approaching the maximum possible cumulative reward, similar to the expert (Fig. \ref{fig:results_metrics_learning}). Evidently, DAgger substantially helped to compensate for errors in perceiving the leader with pressure feedback, even though the training dataset represented only 48 minutes of experimental data. However, the large error bars suggest high variability in performance between rollouts, and the median metrics occasionally worsened in some learning iterations, likely due to variability in leader paths and initial conditions. Performance gains plateaued after the 6\textsuperscript{th} DAgger iteration. Overall, while the learner policies improved in approximating the expert and leader-following, we reached a limit in the attainable performance with our current setup.

\begin{figure}
    \centering
    \includegraphics{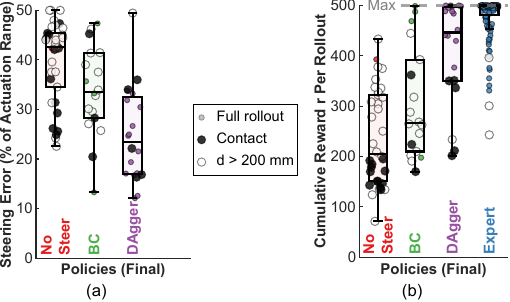}
    \caption{The metric distributions for the policies after training. The boxplots show the median, 25\textsuperscript{th}, and 75\textsuperscript{th} percentiles. Rollouts that ran for the full 10 s are separated from those that ended early due to contact between the µBots or because the distance $d$ (Fig. \ref{fig:paths}) between them became too large. (a) Mean absolute error in steering per rollout between the policy and the expert. (b) Cumulative reward $r$ per rollout.}
    \label{fig:results_metrics_summary}
\end{figure}

The final learner policies rank between the expert and no-steering policies on both metrics (Fig. \ref{fig:results_metrics_summary}). Notably, even the expert performed poorly in some rollouts. These cases highlight the limitations of a steering-only policy, as the expert follower struggled to catch up if it fell behind due to unfavorable early conditions or hydrodynamic interactions. Despite this, the expert consistently achieved the highest cumulative rewards. Both BC and especially DAgger led to several rollouts with near-maximum cumulative rewards, though the inconsistency reflects the limitations of our sensing hardware and policies, even with the leader taking relatively straightforward paths. The low-performing DAgger rollouts were usually due to the follower failing to detect the closing distance and contacting the leader, rather than moving too far away, contrasting with the BC rollouts (Fig. \ref{fig:results_metrics_summary}b). Nonetheless, these results confirm that leader-follower formation swimming can be effectively mediated by flow feedback in robotic and biological fish.

\begin{figure*}[t]
    \centering
    \includegraphics{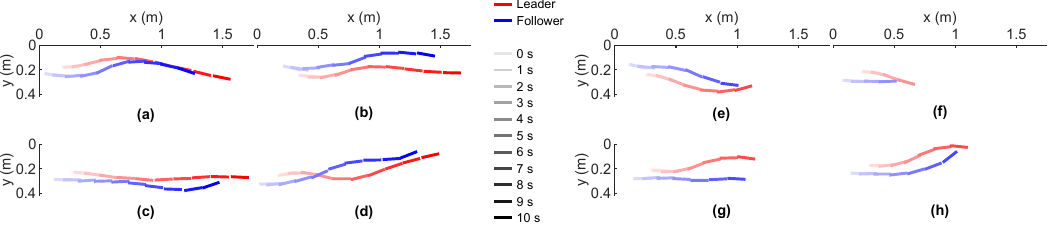}
    \caption{Some paths taken by the leader and the follower using the DAgger-learned policy. The nominal speed was 155\textpm 30 mm/s (0.8\textpm 0.2 body lengths/s).}
    \label{fig:trajectories}
\end{figure*}

Some successful trajectories from the DAgger-learned policy are shown in Fig. \ref{fig:trajectories}a-d (as well as in the attached video) as examples. Unlike the expert which stayed at a fixed offset from the leader, the learner repeatedly alternated between moving toward and away from the leader in attempts to maintain a staggered configuration. Notably, in the final 3 seconds (Fig. \ref{fig:trajectories}a), the follower continued tracking the leader even while directly behind it, where pressure signals were likely reduced (Fig. \ref{fig:fixed_follower}). In Fig. \ref{fig:trajectories}d, the follower crossed from the right to the left of the leader but was able to correct itself and stay on the leader's left. This is remarkable, as the expert policy was only designed to stay on the side it started, but the learner policy adapted to its initial overshoot.

\begin{figure}
    \centering
    \includegraphics[width=\linewidth]{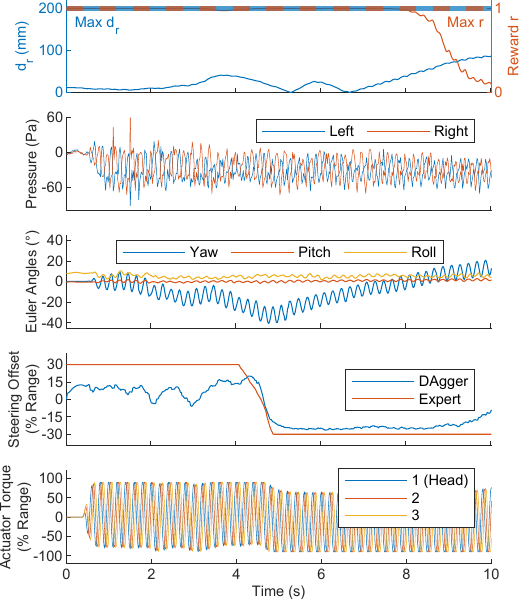}
    \caption{Time-series of $d_{r}$ (Fig. \ref{fig:reward_spatial}), corresponding rewards $r$ (Fig. \ref{fig:reward_dist}), sensor data, steering offset (limited to \textpm30\% of actuation range), and actuator torque commands for the rollout in Fig. \ref{fig:trajectories}a. Positive steering offsets and actuator torques turn left; negative offsets and torques turn right.}
    \label{fig:time-series}
\end{figure}

Examples of data from the motion tracking system, onboard sensors, and policies are shown in Fig. \ref{fig:time-series} for the rollout in Fig. \ref{fig:trajectories}a. Observing relationships between these data types is not straightforward, highlighting the challenge of flow perception in this task. For example, the pressure signals appear more periodic in the last 3 seconds during in-line swimming, but it’s unclear whether earlier irregularities were caused by the follower's changing steering offset or by the leader. This uncertainty is likely unavoidable, as the pressure feedback received by the fixed follower (Fig. \ref{fig:fixed_follower}) is only about 50\% of the amplitude observed here. Despite this uncertainty, the LSTM model was able to approximate the expert's steering trends to a satisfactory degree.

In some scenarios, the follower either touched the leader or moved too far away (Fig. \ref{fig:trajectories}e-h). In Fig. \ref{fig:trajectories}e, the follower steered toward the leader to close the gap and then attempted to steer away, but could not steer or slow down enough to avoid a collision. In Fig. \ref{fig:trajectories}f, the leader made an early turn towards the follower, but the follower failed to detect and react to the flow, possibly because the leader's flow and thrust weakened during its steering motion (Sec. \ref{sec:gaits}), making it harder for the follower to use its pressure signals. In Fig. \ref{fig:trajectories}g, the follower lost track of the leader after steering away slightly while trying to maintain distance. Finally, in Fig. \ref{fig:trajectories}h, the follower steered toward the leader but failed to sense the leader's change in direction, resulting in another collision.

% \subsection{Accuracy of Follower's Perception of Leader}
% \subsubsection{Effects of Leader's Position, Heading, and Phase}
% \subsubsection{Effects of Follower's Self-Generated Flow and Efference Copy}
% \subsection{Task: Starting from Rest to Detect and Track the Leader}

%%%%%%%%%%%%%%%%%%%%%%%%% Conclusions %%%%%%%%%%%%%%%%%%%%%%%%%
\section{Discussion and Future Work}
In this work, we successfully achieved leader-follower (staggered) formation swimming in µBots using an imitation learning framework, leveraging potentially stronger hydrodynamic interactions for perception.
Interestingly, staggered formations have also been shown to improve hydrodynamic efficiency in fish schools \cite{Verma2018}. The relative strength of lateral flows in fish swimming\cite{tytell_hydrodynamics_2004}, contrasting with the stronger axial flows in propeller-based vehicles, may enhance both perception and efficiency in staggered formations. 
%However, decoding pressure feedback in undulating robots, even in such staggered formations, remains challenging due to various noise sources, as indicated by the signals in Fig. \ref{fig:time-series}. 
The follower's successful tracking of the leader's trajectories using this complex feedback demonstrates the effectiveness of recurrent neural networks in extracting relevant information from just two pressure sensors \cite{Verma2018, panta_touchless_2024, Wolf2019}.
% Cite articles on stronger lateral flow in fish swimming and axial flows in propellers.

Despite the progress, challenges remain in improving perception, control, and understanding of hydrodynamic feedback. Adding more pressure sensors (ongoing work) could enhance perception robustness, enabling the follower to better estimate the leader's states across varying leader paths and initial conditions \cite{Zheng2021a}. Extending the policy's degrees of freedom to include the undulation frequency \cite{deng_robot_2024}, amplitude \cite{akanyeti_fish_2016, windsor_swimming_2008, ashraf_burst-and-coast_2020}, and phase offset \cite{Li2020} could also improve both maneuverability and perception, helping avoid failure modes like those in Fig. \ref{fig:trajectories}e-h. We also believe reinforcement learning is the next step in incorporating more sensors and control variables to maximize the utility of hydrodynamic feedback. With the reward signal defined in Fig. \ref{fig:reward}, the follower could learn to identify regions with stronger flow feedback and optimize its gait for better propulsion and perception. Finally, model explanation techniques like SHAP \cite{Lundberg} could help identify key pressure cues used by the imitation-learned policy \cite{panta_touchless_2024}, while particle image velocimetry (PIV) could provide insight into the flow dynamics that generate the pressure feedback \cite{Li2020}.

%%%%%%%%%%%%%%%%%%%%%%%%% References %%%%%%%%%%%%%%%%%%%%%%%%%
% \clearpage
\bibliographystyle{ieeetr}
\bibliography{library}

\end{document}